# Rgtsvm: Support Vector Machines on a GPU in R

**Zhong Wang**  zw355@cornell.edu
**Tinyi Chu**  tc532@cornell.edu
**Lauren A. Choate**  lac334@cornell.edu
**Charles G. Danko**  dankoc@gmail.com
*Baker Institute for Animal Health*
*Cornell University*
*Ithaca, NY, USA*

**Abstract**

*Rgtsvm* provides a fast and flexible support vector machine (SVM) implementation for the R language. The distinguishing feature of *Rgtsvm* is that support vector classification and support vector regression tasks are implemented on a graphical processing unit (GPU), allowing the libraries to scale to millions of examples with >100-fold improvement in performance over existing implementations. Nevertheless, *Rgtsvm* retains feature parity and has an interface that is compatible with the popular *e1071* SVM package in R. Altogether, *Rgtsvm* enables large SVM models to be created by both experienced and novice practitioners.

## 1. Introduction

Support vector machines (SVM) are a powerful family of supervised learning method that provide state-of-art performance on both classification and regression tasks (Cortes and Vapnik, 1995; Drucker et al, 1997). In the classification setting, SVMs learn a hyperplane maximizing the margin that separates training examples drawn from different groups. SVMs also perform well in a regression setting, known as epsilon support vector regression ($\varepsilon$-SVR), in which a predicted value at each training point deviates from the observed response by no more than a tuning parameter, $\varepsilon$. SVMs have a number of appealing properties: they perform well with minimal tuning, generalize reasonably well to different datasets, and can be solved by efficient learning algorithms due to the convex nature of the optimization problem (Cortes and Vapnik, 1995; Drucker et al, 1997; Joachims, 1998). These properties have made SVMs a popular method that is used widely in image processing (Osuna et al, 1997; Plaza et al, 2009; Pradhan 2013; Ma and Guo, 2014), bioinformatics (Danko et al, 2015, Zhang et al 2006, Nassif et al 2009), meteorology (Koprowski, 2013), and other fields.

Of the existing SVM implementations, *libsvm* in C++ performs well on medium-sized learning tasks (Chang and Lin 2011). The SVM feature of the *e1071* (Hornik et al 2006) package offers an R interface to the *libsvm* implementation, which has been widely used by the community thanks to its accessibility, reliability, and ease of use. The primary limitation of *libsvm* is its computing speed and scalability on large datasets, which limits its application to datasets no larger than around 10K training examples. More recently, SVM implementations on general purpose graphical processing units (GPUs) have provided substantial benefits in performance and scalability. The GPU-based SVM implementation *GTSVM* (Cotter et al 2011) takes full advantage of the GPU architecture using CUDA libraries, thereby substantially outperforming other open-sourced SVM implementations. *GTSVM* implements two SVM functions, binary classification and multi-class classification in C/C++. The primary drawback of *GTSVM* is that it the lack of features in more mature and widely used implementations such as *libsvm*. Not only does it lack support vector regression, one of the most commonly used regression methods, but other routine machine learning practices such as hyper-parameter tuning and cross validation are also missing. Furthermore, *GTSVM* requires writing software in C++, which limits its application in rapidly prototyping models and increases the time required to plug new models into machine learning pipelines.

Here we introduce *Rgtsvm*, an open source R package which provides a feature-rich SVM implementation on a GPU. *Rgtsvm* provides >100-fold improvements in performance over *e1071*, while maintaining an identical user interface and features. Users can easily transfer the *svm* function calls in *e1071* to *Rgtsvm* without modification. In addition, *Rgtsvm* expands the features in *GTSVM*, including a fast implementation of epsilon support vector regression on the GPU. Finally, to enable training on large datasets, *Rgtsvm* is written to efficiently manage sparse matrices using the reference class in R, enabling users to easily train SVMs on datasets comprising millions of training examples. *Rgtsvm* is publicly available under GNU General Public License (GPLv3): https://github.com/Danko-Lab/Rgtsvm.

## 2. Installation and Documentation

*Rgtsvm* is an add-on package for R that works on UNIX compatible operating systems. The package depends on CUDA libraries, the *Boost* library, and the *bit64* R package. *Rgtsvm* requires that paths to the *CUDA* and *Boost* libraries are specified by the user during installation, either by an R command or "*install.packages*" in the R console. A vignette containing extensive documentation and examples is provided along with the package. The manual pages provide basic technical information on all functions.

## 3. Implemented Functionality

The aim of *Rgtsvm* is to make a general purpose GPU-based SVM implementation available from R so that large models can be implemented and tested quickly by both novice and experienced users. *Rgtsvm* is implemented as R bindings to our own C/C++ SVM implementation. We based our SVM implementation on the *GTSVM* libraries for an NVIDIA GPU (Cotter et al 2011), and generalized its dual space optimization to enable ε-regression. Additionally, hyper-parameter tuning and cross-validation features are available as function arguments in the R interface. The following sections describe the major features of *Rgtsvm*.

**Binary / multi-class support vector classification:**

Binary and multi-class support vector classification (SVC) is aimed at optimizing a hyperplane that maximizes the decision boundary between two or more training sets. This function is supported in *Rgtsvm* as an R wrapper. We use the same algorithm for optimization that was introduced in *GTSVM*, which takes advantage of the massive parallelism of the GPU. The algorithm works by iteratively optimizing 16 heuristically selected dual space coefficients, known as the working set, until convergence. Each iteration starts by calculating the gradient for all dual space coefficients, followed by picking 16 dual space coefficients based on which partial derivatives of the dual objective function are the largest, subject to dual space constraints. The 16 dual space coefficients are then optimized based on the local gradient. This iteration continues until the primal and dual objective converges. The GPU implementation enables parallel computation of the gradients, which greatly speeds up the optimization.

**Epsilon support vector regression:**

Epsilon support vector regression (ε-SVR) is aimed at learning a regression function that tolerates at most ε-deviation from the true response. Whereas *GTSVM* can only optimize SVC, a generalized optimization scheme is required for ε-SVR. Our implementation is described in this section.

The optimization scheme for SVC can be extended to accommodate ε-SVR. Intuitively the optimization procedure for ε-SVR can be understood as optimizing the dual space objective function over two copies of original training samples introducing an additional label *y* which takes values of +1 or -1, hereafter referred to as the positive and negative copy, respectively. Additionally, the positive and negative copy have their own set of dual space coefficients $\alpha^*$ and $\alpha$ respectively. The dual space objective function of ε-SVR, is then formulated as Eq. 1 (Chang and Lin, 2011), with a similar form as that of SVC.

$$\min_{\alpha,\alpha^*}(\frac{1}{2}[(\alpha^*)^T, \quad \alpha^T]\begin{bmatrix} Q & -Q \\ -Q & Q \end{bmatrix}\begin{bmatrix} \alpha^* \\ \alpha \end{bmatrix} + [\varepsilon E^T - z^T, \quad \varepsilon E^T + z^T]\begin{bmatrix} \alpha \\ \alpha^* \end{bmatrix}) \qquad (1)$$

$$\text{subject to} \quad y^T \begin{bmatrix} \alpha^* \\ \alpha \end{bmatrix} = 0, \qquad 0 \leq \alpha_i, a_i^* \leq C, i = 1, \ldots, l,$$

$$\text{where} \quad y = [\underbrace{1,\ldots,1}_{l}, \quad \underbrace{-1,\ldots,-1}_{l}]^T.$$

Note that α* and α are vectors composed of the dual space coefficients of the positive and negative samples respectively; $Q$ is a $l$ by $l$ positive semidefinite matrix, where $Q_{ij} \equiv y_i y_j K(x_i, x_j)$, and $K(x_i, x_j) \equiv \Phi(x_i)^T \Phi(x_j)$ is the kernel function; $E$ is an all-one vector; $z$ is the vector of the true response; ε is the tolerance; the vector $y$ is composed of $y_i$, which equals +1 for the positive copy and -1 for the negative copy.

The general form of the dual space equation (Eq. 2) is defined as,

$$\min_{\alpha}(\frac{1}{2}\alpha^T Q \alpha + p^T \alpha) \qquad (2)$$

$$\text{subject to} \quad y^T \alpha = 0, \qquad 0 \leq \alpha_i \leq C, i = 1, \ldots, l,$$

By comparing (1) with (2), we found that (1) can be obtained by substituting the generalized linear term $p$ in (2) with $[\varepsilon E^T - z^T, \quad \varepsilon E^T + z^T]$. In binary classification implemented in *GTSVM*, $p^T$ is a linear term of an all-one vector. Therefore, by generalizing the linear term in *GTSVM* to $[\varepsilon E^T - z^T, \quad \varepsilon E^T + z^T]$ during the optimization, the algorithm used for SVC in GTSVM accommodated ε-SVR. The optimization then proceeds in the same way as in binary classification, in which the working set of 16 examples is selected from those with the largest partial derivatives, $\alpha$ and $\alpha^*$ increments are optimized under the constraints, and the responses terms are updated.

**The R interface**

The R interface in *Rgtsvm* implements all of the features of *e1071*. We adopt the same R interface for users to interact with SVC and SVR in order to maintain compatibility with scripts written for the *e1071* implementation. Features provided by the *Rgtsvm* package include those listed below.

    1)     Binary classification, multiclass classification and ε-regression
    2)     4 kernel functions (linear, polynomial, radial basis function and sigmoidal kernel)
    3)     K-fold cross validation
    4)     Tuning parameters in kernel function or in SVM primary space or dual space (C, ε)
    5)     Scaling training data to get better performance.
    6)     Calculating accuracy or correlation, mean square error for the model training.

**Efficient memory handling**

The R environment has intrinsic limitations of R in working with large data sets. By default, R copies all variables passed to function calls, and is therefore prone to memory exhaustion when passing extremely large matrices containing training data to fit a new SVM. Using pointers to avoid unnecessary copies can elegantly solve this issue, yet it is cumbersome for R users and requires other specialized packages, e.g. *Bigmemory* (Kane et al, 2010), and *ff* package (Adler et al, 2008). *Rgtsvm* uses reference classes to encapsulate training variables, such that only their memory address but not the variables themselves are copied when data is passed to the backend C/C++ functions. This strategy optimized memory usage on the CPU side, only leaving the limit on the physical memory of GPU as the only factor limiting the size of training and test matrices passed to *Rgtsvm*.

## 4. Comparison to Similar Toolkits/Frameworks

To the best of our knowledge, *Rgtsvm* is the only GPU-based SVM implementation in the R environment with an open source software license. *RPUSVM* (Liu 2014) is an R package that works on a GPU. However, this package is closed source, requires a paid license, and there is no documentation or publication that explains its algorithm. Several existing GPU-based implementations offer efficient support vector classification, including *GPUSVM* (Li et al 2011), *cuSVM* (Carpenter 2009), *SP-SVM* (Shen et al 2015), and *GTSVM*. None of these packages offer an R interface that allows rapid prototyping of models. Compared to these packages, a major advantage of *Rgtsvm* is its use of the R computing environment, which offers convenient access to the many other statistical and visualization functions included within R.

## 5. Benchmark

The GPU-based architecture in *Rgtsvm* significantly decreases run times compared with *e1071*, the SVM implementation that is most similar in terms of features and accessibility. *Rgtsvm* can optimize 16 dual space coefficients during each iteration and parallelizes kernel functions using GPU threads, greatly reducing the computational time. **Figure 1** demonstrates the difference in performance between *e1071* and *Rgtsvm* implementations. Our *Rgtsvm* package speeds up over 120 times on GPU nodes (NVIDA Tesla K20 GPU) compared to the *e1071* package on CPU nodes (Intel(R) Xeon(R) CPU E5-4620) using 113K samples with 120 features for SVR and 69K samples with 500 features for SVC.

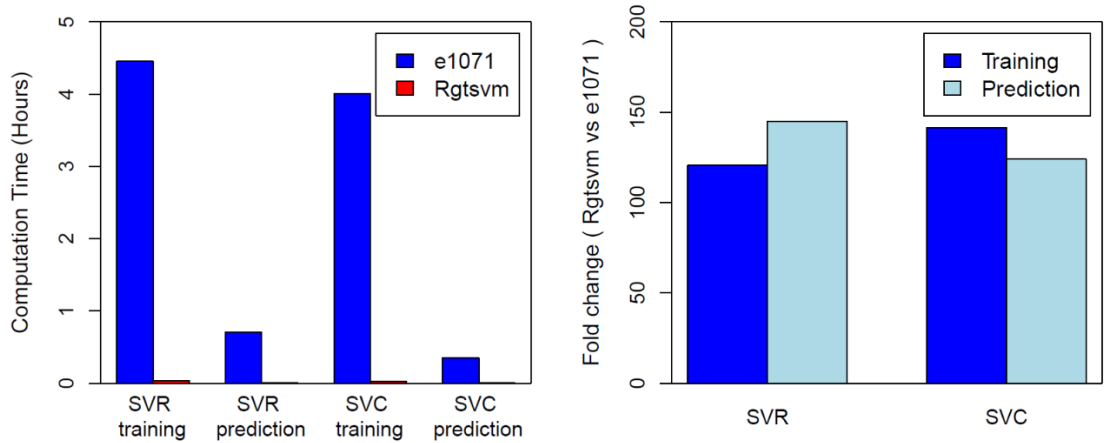

**Fig. 1: Comparison of training and classification time between *e1071* and *Rgtsvm* SVM implementations. (a)** Computation time (hours) for training and predicting a SVR model on 113,978 samples with 120 feature vectors, or a SVC model with 68,971 samples with 501 feature vectors. Training was conducted using *e1071* (CPU implementation based on libsvm) and *Rgtsvm* (GPU implementation). Computations were performed on a Tesla K20 GPU and an E5-4620 Intel Xenon CPU. **(b)** Fold improvement in computation time between *Rgtsvm* and *e1071* for each task described in panel (a).

## Acknowledgments:

This work was supported by National Institutes of Health (NIH) and the National Human Genome Research Institute grant R01-HG009309-01 to C.G.D. The content is solely the responsibility of the authors and does not necessarily represent the official views of the US National Institutes of Health. Computational resources were provided by the Extreme Science and Engineering Discovery Environment (XSEDE) grant TG-MCB160061 and BIO160047 to C.G.D and Z.W.

## References:

Adler, D., Nenadic, O., Zucchini, W., & Glaser, C. (2008). The ff package: Handling Large Data Sets in R with Memory Mapped Pages of Binary Flat Files.

Catanzaro, B., Sundaram, N., & Keutzer, K. (2008, July). Fast support vector machine training and classification on graphics processors. In *Proceedings of the 25th international conference on Machine learning* (pp. 104-111). ACM.

Carpenter, A. U. S. T. I. N. (2009). CUSVM: A CUDA implementation of support vector classification and regression. *patternsonscreen. net/cuSVMDesc. pdf*.

Chang, C. C., & Lin, C. J. (2011). LIBSVM: a library for support vector machines. *ACM Transactions on Intelligent Systems and Technology (TIST)*, *2*(3), 27.

Cortes, C., & Vapnik, V. (1995). Support-vector networks. *Machine learning*, *20*(3), 273-297.

Cotter, A., Srebro, N., & Keshet, J. (2011, August). A GPU-tailored approach for training kernelized SVMs. In *Proceedings of the 17th ACM SIGKDD international conference on knowledge discovery and data mining* (pp. 805-813). ACM.

Danko, C. G., Hyland, S. L., Core, L. J., Martins, A. L., Waters, C. T., Lee, H. W., ... & Siepel, A. (2015). Identification of active transcriptional regulatory elements from GRO-seq data. *Nature methods*, *12*(5), 433-438.

Drucker, H., Burges, C. J., Kaufman, L., Smola, A. J., & Vapnik, V. (1997). Support vector regression machines. In *Advances in neural information processing systems* (pp. 155-161).

Hornik, K., Meyer, D., & Karatzoglou, A. (2006). Support vector machines in R. *Journal of statistical software*, *15*(9), 1-28.

Kane, M. J., Emerson, J. W., & Haverty, P. (2010). bigmemory: Manage massive matrices with shared memory and memory-mapped files. *R package version*, *4*(3).

Koprowski, M. (2013). Spatial distribution of introduced Norway spruce growth in lowland Poland: the influence of changing climate and extreme weather events. *Quaternary International*, 283, 139-146.

Li, Q., Salman, R., Test, E., Strack, R., & Kecman, V. (2011). GPUSVM: a comprehensive CUDA based support vector machine package. *Central European Journal of Computer Science*, *1*(4), 387-405.

Liu, Y. (2014). *GPU-accelerated Computation for Statistical Analysis of the Next-Generation Sequencing Data* (Doctoral dissertation, WORCESTER POLYTECHNIC INSTITUTE).

Joachims, T. (1998). *Making large-scale SVM learning practical* (No. 1998, 28). Technical Report, SFB 475: Komplexitätsreduktion in Multivariaten Datenstrukturen, Universität Dortmund.

Ma, Y. and Guo, D. (2014). *Support vector machines applications*. New York: Springer.

Nassif, H., Al-Ali, H., Khuri, S., & Keirouz, W. (2009). Prediction of protein-glucose binding sites using support vector machines. *Proteins: Structure, Function, and Bioinformatics*, *77*(1), 121-132.

Osuna, E., Freund, R., & Girosit, F. (1997, June). Training support vector machines: an application to face detection. In *Computer vision and pattern recognition, 1997. Proceedings., 1997 IEEE computer society conference on* (pp. 130-136). IEEE.

Plaza, A., Benediktsson, J. A., Boardman, J. W., Brazile, J., Bruzzone, L., Camps-Valls, G., ... & Marconcini, M. (2009). Recent advances in techniques for hyperspectral image processing. *Remote sensing of environment*, *113*, S110-S122.

Pradhan, B. (2013). A comparative study on the predictive ability of the decision tree, support vector machine and neuro-fuzzy models in landslide susceptibility mapping using GIS. *Computers & Geosciences*, *51*, 350-365.

Shen, B., Liu, B. D., Wang, Q., Fang, Y., & Allebach, J. P. (2015, January). SP-SVM: Large Margin Classifier for Data on Multiple Manifolds. In *AAAI* (pp. 2965-2971).

Zhang, H. H., Ahn, J., Lin, X. and Park, C. (2006). Gene selection using support vector machines with nonconvex penalty. *Bioinformatics*, 22, pp. 88-95